\pdfoutput=1

\documentclass[11pt]{article}
\PassOptionsToPackage{table}{xcolor}

\usepackage[preprint]{acl}

\usepackage{times}
\usepackage{latexsym}

\usepackage[T1]{fontenc}

\usepackage[utf8]{inputenc}

\usepackage{microtype}

\usepackage{inconsolata}

\usepackage{graphicx}
\usepackage{wrapfig}
\usepackage{floatflt}
\usepackage{microtype}
\usepackage{hyperref}
\usepackage{url}
\usepackage{booktabs}

\usepackage{lineno}
\usepackage{xurl}
\usepackage{times}
\usepackage{latexsym}
\usepackage{soul}
\sethlcolor{yellow}  
\usepackage{CJKutf8}
\usepackage{xcolor}    
\usepackage{floatflt}
\usepackage{svg}
\usepackage{amsmath} 
\usepackage{enumitem} 
\usepackage{booktabs}
\usepackage{multirow}
\usepackage{microtype} 
\usepackage{algorithm}
\usepackage{algpseudocode}
\usepackage{amsmath}
\usepackage{tcolorbox}
\usepackage{microtype}

\usepackage{inconsolata}
\usepackage{graphicx}

\usepackage[utf8]{inputenc}
\usepackage{tcolorbox}
\usepackage{hyperref}
\usepackage{amsmath}
\usepackage{xcolor}
\usepackage{listings} 
\usepackage{titlesec} 
\usepackage{tcolorbox}
\usepackage{times}
\usepackage{latexsym}
\usepackage{tabularx}
\usepackage[T1]{fontenc}
\usepackage[utf8]{inputenc}
\usepackage{amsmath,amssymb}
\usepackage{graphicx}
\usepackage{booktabs}
\usepackage{inconsolata}
\usepackage{microtype}
\usepackage{algorithm}
\usepackage{algpseudocode}
\usepackage{url}
\usepackage{multirow}
\usepackage{subcaption}
\usepackage{enumitem}
\usepackage{hyperref}
\usepackage{xcolor}
\usepackage{listings}
\usepackage{titlesec}
\usepackage{longtable}
\usepackage{setspace}
\usepackage[table]{xcolor}

\usepackage{booktabs}      
\usepackage{tabularx}      
\usepackage{caption}       
\usepackage{longtable}     
\usepackage{hyperref}      
\usepackage{setspace}      
\usepackage[T1]{fontenc}

\definecolor{myframecolor}{RGB}{70, 130, 180} 
\definecolor{mybackcolor}{RGB}{240, 248, 255} 

\lstdefinelanguage{json}{
    basicstyle=\ttfamily\small, 
    string=[b]", 
    comment=[l]{:}, 
    morestring=[b]',
    morecomment=[l]{:}, 
    showstringspaces=false, 
    breaklines=true, 
    breakatwhitespace=true, 
}

\tcbset{
    myboxstyle/.style={
        colframe=myframecolor,
        colback=mybackcolor,
        boxrule=0.5mm,
        boxsep=5pt,
        arc=4pt,
        auto outer arc,
        listing engine=listings,      
        listing only,
        listing options={
            breaklines=true,
            breakatwhitespace=true,
            basicstyle=\ttfamily\small,
            language=JSON
        },
        title style={font=\bfseries},
        left=5pt,
        right=5pt,
        top=5pt,
        bottom=5pt
    }
}

\definecolor{myframecolor}{RGB}{66,115,176}
\definecolor{mybackcolor}{RGB}{202,220,238}

\definecolor{darkblue}{rgb}{0, 0, 0.5}
\hypersetup{colorlinks=true, citecolor=darkblue, linkcolor=darkblue, urlcolor=darkblue}

\title{ReAgent: Reversible Multi-Agent Reasoning for \\Knowledge-Enhanced Multi-Hop QA}

\author{%
  \textbf{Xinjie~Zhao\textsuperscript{1}\textsuperscript{\ensuremath{\dagger}}},
  \textbf{Fan~Gao\textsuperscript{1}\textsuperscript{\ensuremath{\dagger}}},
  \textbf{Xingyu~Song\textsuperscript{1}},
  \textbf{Yingjian~Chen\textsuperscript{1}},
  \textbf{Rui~Yang\textsuperscript{2}},
  \textbf{Yanran~Fu\textsuperscript{1}},\\
  \textbf{Yuyang~Wang\textsuperscript{3}},
  \textbf{Yusuke~Iwasawa\textsuperscript{1}},
  \textbf{Yutaka~Matsuo\textsuperscript{1}},
  \textbf{Irene~Li\textsuperscript{1}}\\[6pt]
  \textsuperscript{1}The University of Tokyo\quad
  \textsuperscript{2}Duke–NUS Medical School\quad
  \textsuperscript{3}The Hong Kong University of Science and Technology\\[4pt]
  \small\href{mailto:irene.li@weblab.t.u-tokyo.ac.jp}{irene.li@weblab.t.u-tokyo.ac.jp}
}

\begin{document}
\maketitle

\begingroup
  \renewcommand\thefootnote{\ensuremath{\dagger}}
  \footnotetext{Equal contribution.}
\endgroup

\begin{abstract}

Multi-hop question answering (QA) remains challenging, as solutions must reliably integrate and reconcile evidence from multiple sources without succumbing to error propagation. While large language models (LLMs) have achieved substantial improvements via chain-of-thought (CoT) prompting and retrieval-augmented generation, these methods typically adopt a forward-only workflow—early mistakes persist throughout inference, and contradictions discovered later cannot systematically trigger re-evaluation.
To address this limitation, we present \emph{ReAgent}, a \emph{reversible} multi-agent reasoning framework. Specifically, ReAgent enables agents to backtrack to earlier valid states when conflicts arise, thereby isolating and rectifying flawed assumptions before they undermine subsequent reasoning. Our approach combines explicit local and global rollback protocols with modular role specialization, resulting in a flexible and error-tolerant pipeline. Empirical evaluation on three multi-hop QA benchmarks demonstrates consistent performance gains of approximately 6\% over forward-only baselines, in addition to enhanced interpretability. These findings highlight the value of non-monotonic, backtracking-driven inference in complex QA scenarios and point to broader implications for multi-agent collaboration in knowledge-intensive tasks.\footnote{Our anonymous code is available at \url{https://anonymous.4open.science/r/ReAgent-9415}.}
\end{abstract}

\section{Introduction}
Multi-hop question answering (QA) is a central challenge in natural language processing (NLP), demanding the ability to gather and integrate evidence across multiple documents, database entries, or knowledge-graph nodes before converging on a single correct answer \citep{yang2018hotpotqa, welbl2018constructing, ho-etal-2020-constructing}. Benchmarks such as HotpotQA \citep{yang2018hotpotqa} and 2WikiMultiHopQA \citep{ho-etal-2020-constructing} highlight the intricate nature of multi-hop inference, where each reasoning step can involve partial retrieval, validation, and synthesis of new information. A core difficulty lies in the system's vulnerability to early mistakes: if an incorrect inference is made at an initial hop, subsequent steps often propagate this error and undermine the final outcome, rendering it contradictory or simply wrong \citep{inoue-etal-2020-r4c, bo2024copper,yang2024graphusion}. This phenomenon has motivated extensive research into chaining intermediate inferences and exploring ways to detect, isolate, or rectify problematic conclusions.
\begin{figure}[t]
    \centering
    \includegraphics[width=0.99\linewidth]{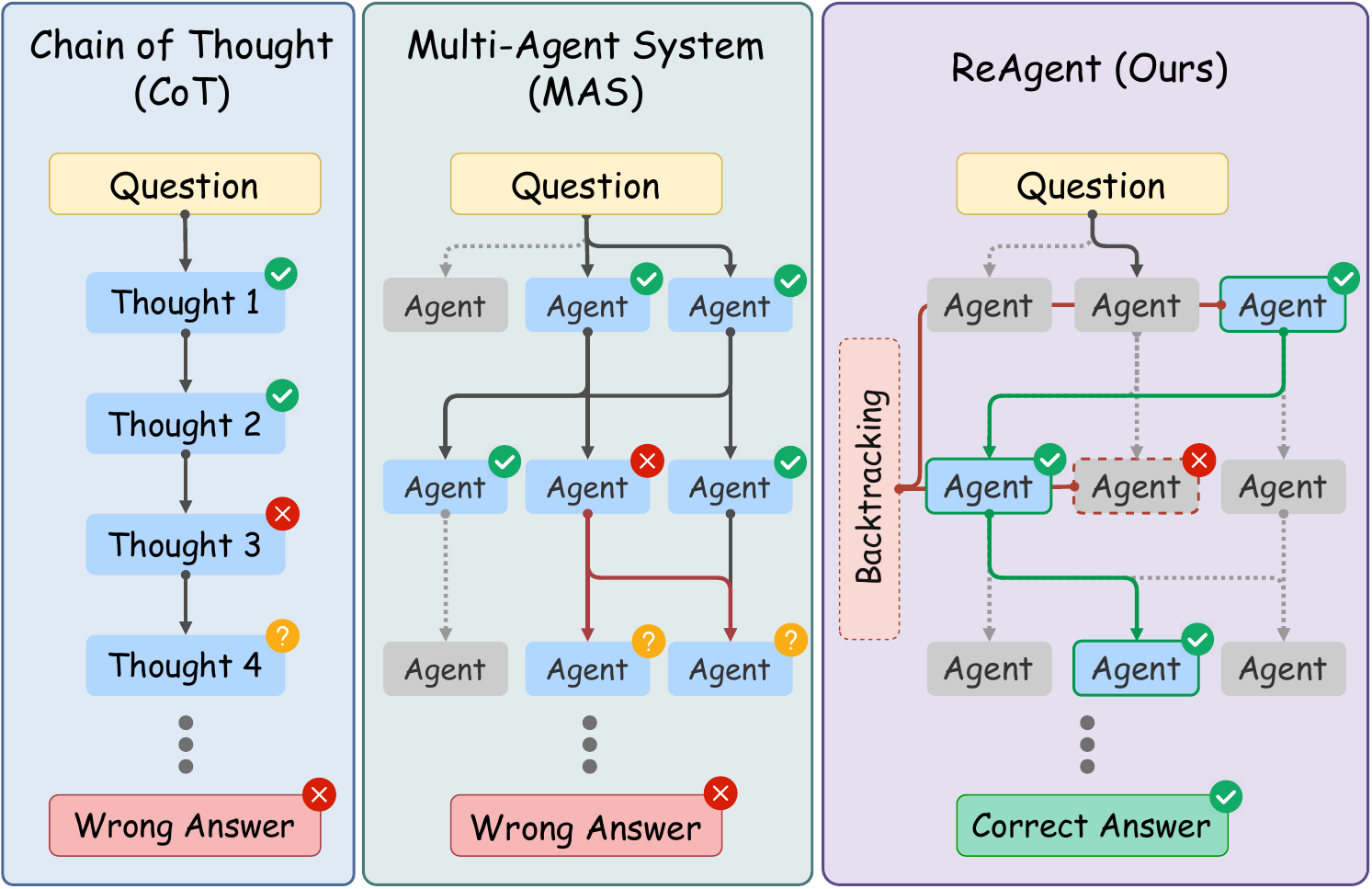}
    \caption{Comparison of multi-hop reasoning strategies. \textbf{Chain-of-Thought (CoT)} and \textbf{Multi-Agent Systems (MAS)} typically adopt a forward-driven reasoning pipeline without rollback mechanisms, which could generate the wrong answer due to error accumulation. In contrast, our proposed \textbf{ReAgent} introduces explicit backtracking mechanisms that enable the system to correct errors during reasoning, resulting in a more accurate and reliable answer.}
    \label{fig:1}
    \vspace{-4mm}
\end{figure}

Recent large language models (LLMs) exhibit promising results on multi-hop QA, frequently using either explicit Chain-of-Thought (CoT) prompting \citep{wang2023selfconsistency} (Figure~\ref{fig:1}, left) or retrieval-augmented generation \citep{das2018minerva,long2019deeprl, Gao2023EvaluatingLL, Liu2023RecPromptAS, yang-etal-2024-kg, yang2025retrieval}. These methods facilitate a stepwise approach, encouraging transparency in how each intermediate fact is reached. Nonetheless, they typically rely on a forward-driven reasoning pipeline that does not proactively examine or correct previously accepted statements. In practice, once a model commits an erroneous partial inference may not reevaluate it unless given targeted prompts to do so \citep{puerto-etal-2023-metaqa, huang2024mirror}. This unidirectional paradigm is problematic when later-discovered evidence or reasoning paths contradict prior assumptions, as the system lacks a structured mechanism to revise and propagate corrections. Although incremental improvements have been proposed, the absence of robust backtracking or rollback still limits their reliability and interpretability \citep{doyle1979tms, bo2024copper}.

A growing body of work in multi-agent collaboration (Figure~\ref{fig:1}, middle) offers an alternative perspective, assigning specialized components distinct roles in retrieval, validation, conflict detection, and results assembly \citep{zhao2024longagent, parhizkar2020trust, info:doi/10.2196/59439}. Approaches such as \textit{COPPER} \citep{bo2024copper} and \textit{LongAgent} \citep{zhao2024longagent} distribute the QA task among multiple LLM-based agents that communicate via message passing, thereby providing greater modularity and a clearer division of labor. By cross-verifying evidence, each agent can potentially identify suspicious partial solutions. However, even these designs often lack a systematic strategy to revert to an earlier, valid state once a global contradiction emerges. Such a reversal capability is non-trivial, as it introduces synchronization complexities among the agents and raises questions about how to detect, prioritize, and handle contradictory or low-confidence information in a large-scale collaborative setting \citep{he2021intermediate}.

In this paper, we propose \textbf{ReAgent}, a \textit{reversible multi-agent collaborative reasoning framework} for multi-hop QA (Figure~\ref{fig:1}, right). Our method introduces explicit backtracking protocols into a multi-layer architecture that addresses both the granular aspects of local error correction and the broader system-wide consistency checks. 
ReAgent alleviates error accumulation in forward-only strategies by incorporating fine-grained conflict-detection cues at each step and using a flexible error-correction loop that can revert and iteratively refine intermediate inferences. Our design is inspired by multi-agent systems with reflective capabilities \citep{bo2024copper, huang2024mirror}, but we go further by defining how local versus global backtracking unfolds, how to manage concurrency issues when parallel agents must revert their states, and how to integrate trust signals derived from each agent’s past reliability \citep{puerto-etal-2023-metaqa, parhizkar2020trust}. While the addition of reversible reasoning raises legitimate questions about computational overhead and concurrency, our findings show that these challenges can be mitigated by careful coordination at the supervisory level.
Experiments on three public multi-hop QA benchmarks demonstrate that ReAgent improves final-answer accuracy while enhancing interpretability, and even outperforms several advanced reasoning models despite it built on a lightweight, non-reasoning foundation.

\begin{figure*}[t]
    \centering
    \vspace{-4mm}
    \includegraphics[width=\textwidth]{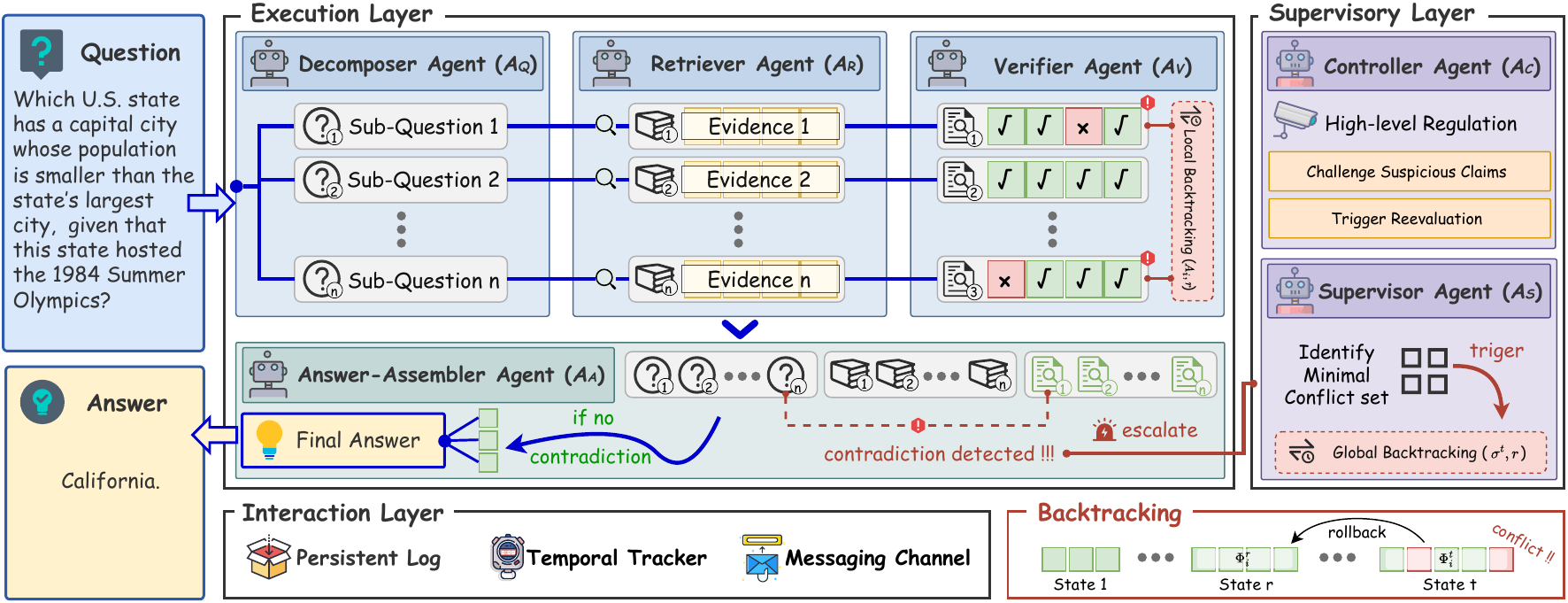}
    \caption{The overall architecture of the ReAgent. The given question is processed through the Execution Layer, which involves question decomposition, evidence retrieval, verification, and is ultimately integrated to generate the final answer (\textcolor[HTML]{0000CC}{blue line}). The Supervisor Layer and Interaction Layer are responsible for monitoring, regulation, and communication. The ReAgent framework includes both local and global backtracking mechanisms (\textcolor[HTML]{AE4132}{red boxes}), triggered by the Verifier Agent ($A_V$) and Supervisor Agent ($A_S$), respectively.}
    \label{fig:2}
    \vspace{-4mm}
\end{figure*}

Our contributions are threefold: (1) we propose a multi-agent QA framework that supports both local and global backtracking to correct mistakes in situ; (2) we design a hybrid retrieval mechanism that integrates textual and graph-based evidence; and (3) We empirically demonstrate the effectiveness of ReAgent, achieving an average accuracy gain of approximately 6\% over the strongest baselines, while improving robustness and transparency compared to forward-only methods.

\section{Related Work}
\textbf{Prompting and Iterative Reasoning.} 
Large Language Models (LLMs) can tackle complex tasks using chain-of-thought (CoT) prompting\cite{yang2024qwen2,yang2025graphusion}, which promotes step-by-step reasoning and improves performance in arithmetic, commonsense, and symbolic tasks \citep{wei2022chain}. Accuracy further improves with self-consistency, which samples multiple reasoning paths and selects the majority answer \citep{wang2023selfconsistency}. To reduce manual prompt design, automatic methods such as APE generate and evaluate prompts via LLMs themselves, treating prompt creation as a search task \citep{zhou2023large}. Iterative frameworks like ReAct combine reasoning with tool use to refine answers \citep{yao2023react}, while Reflexion and Self-Refine add feedback loops where the model critiques and revises its outputs \citep{shinn2023reflexion, madaan2023selfrefine}. Prompting has also extended to multimodal inputs; for instance, Multimodal-CoT integrates visual and textual information into a joint reasoning trace to generate the final answer \citep{zhang2023multimodal}.

\noindent\textbf{Multi-Agent Collaboration, Debate, and Scalability.}
Recent work investigates multi-agent systems where multiple LLMs collaborate or compete toward shared goals. CAMEL assigns roles (e.g., “user” and “assistant”) to agents that communicate via dialogue to decompose tasks \citep{li2023camel}. Role specialization and consensus mechanisms enhance robustness. Debate-based methods push this further, with agents arguing opposing views and a judge—human or model—selecting the best solution \citep{khan2024debate, xiong2023interconsistency}. Such adversarial exchanges promote correctness, though dominant agents can bias results without capable judges. As agent counts grow, coordination becomes a bottleneck. Efficient structures like hierarchical controllers and learned protocols are needed to manage scalability and communication overhead \citep{li2023camel, xiong2023interconsistency}.

\section{Preliminary}
\label{sec:preliminary}
\noindent\textbf{Multi-Hop QA Setup.}
Multi-hop QA tasks aim to answer a query $Q$ by integrating evidence $\mathcal{E}$ from multiple sources through a series of reasoning steps, where each hop contributes partial information toward the final answer. In a typical forward-only pipeline, the process can be formalized as:
\(
    Q \longmapsto \{\,e_1, e_2, \ldots, e_k\}
    \longmapsto\text{inferred statement(s)}
    \longmapsto\text{Final Answer}
\), where $\{e_1, e_2, \ldots, e_k\} \subset \mathcal{E}$ represents a potentially large pool of evidence, each \(e_i\) denotes evidence used at the \(i\)-th step of the reasoning chain.

\noindent\textbf{Non-Monotonic Backtracking.}
Our notion of \emph{backtracking} is a non-monotonic extension of the typical multi-hop QA process. We introduce a reversible reasoning mechanism that allows both local and global backtracking. 
\begin{itemize}[noitemsep, topsep=1pt, leftmargin=1em]
    \item \textbf{Local Correction}: An agent can revise its own inference when it detects internal conflict or receives contradictory evidence from other agents.
    \item \textbf{Global Rollback}: A supervisor coordinates rollback across agents when inconsistencies span multiple agents, restoring the system to a previously consistent state.
\end{itemize}

\section{ReAgent: Reversible Multi-Agent Reasoning Architecture}
\label{sec:approach}

Figure~\ref{fig:2} presents the overall architecture of our proposed ReAgent, organized into three layers: 1) \textbf{Execution Layer}, responsible for decomposing the input question $Q$ into multiple sub-questions, retrieving relevant evidence respectively, validating intermediate results, and integrating them to generate the final answer $A_{final}$; 2) \textbf{Supervisor Layer}, responsible for high-level regulation, coordinating conflict resolution and managing global backtracking; and 3) \textbf{Interaction Layer}, responsible for maintaining the concurrency model and communication protocols.

The core of the architecture is a \textit{hierarchical backtracking mechanism}, consisting of local backtracking, which resolves internal contradictions within each agent, and global backtracking, which handles contradictions spanning multiple agents. To support this, ReAgent maintains \textit{knowledge sets} at each time step $t$. Specifically, given a set of agents $\mathcal{A} = \{A_1, A_2, \dots, A_n\}$, each agent $A_i$ holds a local knowledge set $\Phi_i^t$, containing its proposed assertions, retrieved evidence, or intermediate inferences, while the system maintains a global knowledge set $\Phi^t = \bigcup_{i=1}^{n} \Phi_i^t$, representing the set of global statements under consideration at time $t$. The design explicitly supports \emph{non-monotonic} updates: newly introduced statements can be revoked if they lead to logical conflicts or are superseded by contradictory evidence. The specific prompts of each kind of agent are provided in Appendix~\ref{sec:appendix-prompts}.

\subsection{Execution Layer Agents}
\label{subsec:base_layer}

The Execution Layer hosts four types of agents that address fundamental QA sub-tasks, each maintaining its local knowledge $\Phi_i$. 


\noindent\textbf{Question-Decomposer Agent ($A_Q$).}
Given a complex input question $Q$, this agent breaks it into a set of sub-questions for subsequent retrieval and verification: $A_Q: Q \longmapsto \{q_1, \dots, q_m\}$. The decomposition is broadcast to other base-layer agents.

\noindent\textbf{Retriever Agent ($A_R$).}
Upon receiving a sub-question $q_i$, $A_R$ issues parallel sparse and dense queries over the corpus, merges the hits with reciprocal-rank fusion, and retains the top-$M$ passages as the evidence set $E={e_1,\dots,e_M}$. If backtracking invalidates any $e_j \in E$, only the associated $q_i$ is re-queried, leaving the rest unchanged.

\noindent\textbf{Verifier Agent ($A_V$).}
Performs local consistency checks. Given a new evidence set $E$, it verifies coherence with its current knowledge $\Phi_V^t$. If conflicts arise, it triggers local backtracking to revert to a consistent state or signals higher-level agents for broader resolution. Concretely, $A_V$ may invoke $\mathrm{BacktrackLocal}\bigl(A_V, r\bigr)$
if it detects inconsistencies in assertions introduced after node $r$.

\noindent\textbf{Answer-Assembler Agent ($A_A$).}
Gathers partial answers (and verified evidence) to synthesize a final answer. Given the local inferences from $A_Q$, $A_R$, and $A_V$, the agent combines them to generate the final answer $A_{final}$, represented as: 
\begin{equation*} 
    A_{final} = A_A(\Phi_Q, \Phi_R, \Phi_V),
\end{equation*}
where $\Phi_Q$, $\Phi_R$ and $\Phi_V$ denotes the knowledge set from the respective agents. Any detected contradiction spanning multiple agents triggers escalation to the supervisory layer.

\subsection{Supervisory Layer Agents}
\label{subsec:supervisory_layer}
The supervisory layer oversees system-wide strategies, especially when contradictory goals or inconsistent states appear across multiple agents that cannot be resolved by a single agent itself.

\noindent\textbf{Controller Agent ($A_C$).}
Regulates high-level strategies by monitoring game-theoretic signals or meta-rules. If a certain rule is deemed suboptimal or unsafe, $A_C$ can override it or enforce mode switches. For instance, it may issue a $\textsf{challenge}(\varphi)$ to examine a crucial assertion $\varphi$ from multiple agents’ perspectives or override local decisions if they jeopardize overall consistency.

\noindent\textbf{Supervisor Agent ($A_S$).}
Coordinates multi-agent conflicts spanning multiple knowledge sets or critical shared assumptions. If a conflict persists after local backtracking and escalation from $A_A$, the agent determines whether partial or holistic rollback is required. Specifically, when a system-wide contradiction is found, $A_S$ identifies a minimal conflict set $\Psi$ such that $\mathrm{SAT}(\Psi) = \textsf{false}$ but $\mathrm{SAT}(\Psi') = \textsf{true}$ for any proper subset $\Psi' \subset \Psi$. The supervisor then triggers a backtracking operation that eliminates or modifies $\Psi$.

\subsection{Interaction Layer}
\label{subsec:interaction_layer}
The Interaction Layer is responsible for storing knowledge sets from each interaction round, as well as maintaining the concurrency model and communication protocols.

\noindent\textbf{Persistent Log.}
Stores the local knowledge sets $\Phi_i$ and global knowledge set $\Phi$ from all interaction rounds to support backtracking and serve as a historical evidence repository.

\noindent\textbf{Temporal Tracker.}
Records the chronological sequence of messages and actions, enabling agents to reference previous steps accurately. Specifically, the system can use temporal operators such as $\square \varphi$ to denote that $\varphi$ must hold in every future state, and $\lozenge \varphi$ to indicate that a proposition $\varphi$ might become true at some future point. These temporal constraints assist in specifying persistent axioms or potential triggers for backtracking conditions.

\noindent\textbf{Messaging Channel.}
Achieves communication across agents for exchanging updates, signaling conflicts, and broadcasting intermediate conclusions. Specifically, atomic events $\mathtt{msg}(\varphi)$ are typed as \textsf{assert}, \textsf{inform}, \textsf{reject}, or \textsf{challenge}. These updates can occur simultaneously, and a composite event model merges parallel messages. Furthermore, a shared concurrency mechanism processes simultaneous actions, ensuring that conflicts arising from concurrent assertions are escalated to the supervisory layer.

\subsection{Conflict Management and Backtracking}

This part formalizes how the system detects contradictions and reverts to consistent states. At each time $t$, the set of all accepted assertions is $\Phi^t$. A conflict occurs if $\Phi^t$ entails both $\phi$ and $\neg\phi$ for some proposition $\phi$, meaning $\mathrm{SAT}(\Phi^t)$ fails. To resolve the issue, the system follows a two-level approach:

\noindent\textbf{Local Backtracking.}
Each agent $A_i$ maintains a backtracking graph $\mathrm{LBG}_i$, which logs states $\Phi_i^r$ at selected checkpoints $r$. If an internal contradiction $\mathrm{Conflict}_i(\Phi_i^t)$ is detected by $A_V$, local backtracking is performed to revert $A_i$ to a prior consistent node $r < t$:
\begin{equation*}
    \mathrm{BacktrackLocal}(A_i,\,r):
    \Phi_i^t \longrightarrow \Phi_i^r.
\end{equation*}
After the local rollback, the agent re-evaluates newly arriving evidence.

\noindent\textbf{Global Backtracking.}
When contradictions span multiple agents, the $A_S$ identifies a minimum conflict set of assertions that must be revised. A global backtracking operation: $\mathrm{BacktrackGlobal}\bigl(\sigma^t,\,r\bigr)$ reverts \emph{all} agents from time $t$ to $r$. The system discards any statements introduced between $r+1$ and $t$, restoring consistency. If the conflict cannot be removed even after a global rollback, the Controller might enforce strategic overrides or disclaim an answer. The corresponding algorithm is detailed in Appendix~\ref{appendix:algorithm}.

\section{Experiments}
\label{sec:experiments}
\subsection{Experimental Setup}

We evaluate ReAgent on three widely used, knowledge-intensive multi-hop QA benchmarks: HotpotQA~\citep{yang2018hotpotqa}, 2WikiMultiHopQA~\citep{ho-etal-2020-constructing}, and MuSiQue~\citep{trivedi-etal-2022-musique}. We compare ReAgent against three main groups of baselines: (1) standard LLMs, (2) dedicated reasoning models, and (3) agent-based models. Experimental results demonstrate that ReAgent consistently outperforms all baseline groups and is particularly effective in solving tasks that require iteratively corrected reasoning. 

\noindent\textbf{Datasets.}
\textbf{HotpotQA} is a large-scale, open-domain dataset that explicitly promotes multi-hop reasoning by requiring the integration of information across multiple full-length Wikipedia passages. \textbf{2WikiMultiHopQA} selects distinct Wikipedia domains for each question, focusing on evaluating the model's multi-hop inference over different resources rather than one document. \textbf{Musique} is a challenging dataset requiring model's ability to reason over multiple dispread sentences. Following previous work~\citep{trivedi-etal-2023-interleaving, press-etal-2023-measuring, gutiérrez2024hipporag}, we utilize 1000 random samples from each validation set and corresponding texts as the knowledge base.

\noindent\textbf{Baselines.} We meticulously categorize diverse models into different groups to compare their capabilities in multi-hop QA. (1) \textbf{Regular Models:} We select non-reasoning models in this group. The models include: Llama-4~\footnote{\url{https://ai.meta.com/blog/llama-4-multimodal-intelligence/}}, Qwen-2.5~\citep{yang2024qwen2} series, DeepSeek-V3-2024-03~\citep{liu2024deepseek}, Genmini-1.5-Flash-2024-09~\citep{team2023gemini}, Gemini-2.0-flash-2025-02~\citep{team2023gemini}, GPT-4o-latest~\citep{hurst2024gpt}, GPT-4.1~\citep{meta2024llama4}, GPT-4o with CoT~\citep{wei2023chainofthoughtpromptingelicitsreasoning}. (2) \textbf{Reasoning Models:} This group includes several strong reasoning baselines for comparison. The models include: Qwen-3-Thinking~\citep{yang2025qwen3technicalreport} in 32B and 253B size, DeepSeek-R1~\citep{deepseekai2025deepseekr1incentivizingreasoningcapability}, Gemini-2.5-pro~\citep{google2025gemini2.5}, O1~\citep{openai2024openaio1card}, and O3~\citep{openai2025o3o4mini}. (3) \textbf{Agentic Models.} Chain-of-Agents (CoA)~\citep{zhang2024chainagentslargelanguage}, HippRAG~\citep{gutiérrez2025hipporagneurobiologicallyinspiredlongterm},  KAG~\citep{liang2024kagboostingllmsprofessional}, and our method. To compare fairly, we employ GPT-4o as the backbone for both COT and Agentic Models.

\noindent\textbf{Implementation.} For large-scale LLMs (e.g., DeepSeek-V3, the Gemini family, GPT-4o, and Qwen-3-235B), we access the models via API. For the medium-sized LLMs, we use their official open-source repositories. The temperature is set to $0.3$ to ensure deterministic outputs, while other parameters follow their default settings. ReAgent is implemented using GPT-4o as the backbone model. Specifically, we set the temperature to $0.8$ for the decomposition agent to encourage diverse reasoning paths, and $0.6$ for all other agents. The cost analysis for proprietary models and agentic methods is presented in Appendix~\ref{appendix:cost}. For open-source models, all experiments are conducted using four A100-80G GPUs.


\noindent\textbf{Metrics.}
We measure \textbf{Exact Match (EM)} and \textbf{F1} scores for the QA evaluation.

\begin{table*}[t]
    \vspace{-4mm}
    \centering
    \resizebox{1.0\textwidth}{!}{
    \setlength{\tabcolsep}{11pt}
    \begin{tabular}{l cc|cc|cc|cc}
        \toprule
        \multirow{2}{*}{\textbf{Model}} & \multicolumn{2}{c}{\textbf{HotpotQA}} & \multicolumn{2}{c}{\textbf{2Wiki}} & \multicolumn{2}{c}{\textbf{Musique}} & \multicolumn{2}{c}{\textbf{Average}}\\
        \cmidrule(lr){2-3} \cmidrule(lr){4-5} \cmidrule(lr){6-7} \cmidrule(lr){8-9}
        & EM & \multicolumn{1}{c}{F1} & EM & \multicolumn{1}{c}{F1} & EM & \multicolumn{1}{c}{F1} & EM & F1 \\
        \midrule
        \multicolumn{7}{l}{\textbf{Regular Models}} \\
         Llama-4-Scout-17B-16E-Instruct & 0.263 & 0.389 & 0.332 & 0.467 & 0.107 & 0.185 & 0.234 & 0.346 \\
        DeepSeek-V3 & 0.352 & 0.491 & 0.466 & 0.579 & 0.223 & 0.33 & 0.347 & 0.467 \\
        Qwen-2.5-32B-Instruct & 0.372 & 0.509 & 0.557 & 0.663 & 0.159 & 0.273 & 0.363 & 0.482 \\
        Qwen-2.5-72B-Instruct & 0.363 & 0.519 & 0.543 & 0.631 & 0.222 & 0.327 & 0.376 & 0.492 \\
        Gemini-1.5-Flash & 0.374 & 0.488 & 0.563 & 0.650 & 0.208 & 0.310 & 0.381 & 0.482 \\
        Gemini-2.0-Flash & 0.371 & 0.490 & 0.538 & 0.651 & 0.246 & 0.338 & 0.385 & 0.493 \\
        GPT-4o & 0.381 & 0.549 & 0.517 & 0.649 & 0.245 & 0.379 & 0.381 & 0.525 \\
        GPT-4.1 & 0.389 & \underline{0.563} & 0.544 & \underline{0.665} & 0.271 & \underline{0.413} & 0.401 & \underline{0.547} \\
        CoT (GPT-4o) & \underline{0.408} & 0.531 & \underline{0.558} & 0.638 & \underline{0.272} & 0.360 & \underline{0.413} & 0.509 \\
        \midrule
        \multicolumn{7}{l}{\textbf{Reasoning Models}} \\
        Qwen-3-32B-Thinking & 0.332 & 0.474 & 0.241 & 0.357 & 0.387 & 0.511 & 0.387 & 0.511 \\
        
        DeepSeek-R1 & 0.356 & 0.483 & 0.601 & 0.707 & 0.298 & 0.416 & 0.418 & 0.535 \\
        Qwen-3-235B-A22B-Thinking & 0.361 & 0.506 & 0.624 & 0.729 & 0.271 & 0.387 & 0.418 & 0.540 \\
        Gemini-2.5-Pro & 0.430 & 0.560 & \cellcolor{pink!42}\underline{0.743} & \cellcolor{pink!42}\underline{0.829} & 0.383 & 0.491 & 0.518 & 0.626 \\
        O1 & 0.505 & 0.661 & 0.656 & 0.758 & \cellcolor{cyan!15}0.417 & \cellcolor{cyan!15}0.551 & 0.526 & 0.656 \\
        O3 & \underline{0.535} & \underline{0.696} & 0.706 & 0.787 & \cellcolor{pink!42}\underline{0.442} & \cellcolor{pink!42}\underline{0.579} & \cellcolor{cyan!15}\underline{0.561} & \cellcolor{cyan!15}\underline{0.687} \\
        \midrule
        \multicolumn{7}{l}{\textbf{Agentic Models (w.GPT-4o)}} \\
        CoA~\cite{zhang2024chainagentslargelanguage} 
         & 0.391 & 0.558 & 0.575 & 0.697 & 0.239 & 0.361 & 0.402 & 0.539 \\
        HippoRAG~\citep{gutiérrez2025hipporagneurobiologicallyinspiredlongterm} & 0.528 & 0.717 & 0.633 & 0.725 & 0.353 & 0.507 & 0.504 & 0.649 \\
        KAG~\citep{liang2024kagboostingllmsprofessional}& \cellcolor{cyan!15}0.603 & \cellcolor{cyan!15}0.782 & 0.681 & 0.781 & 0.348 & 0.489 & 0.544 & 0.684\\
        \textbf{ReAgent (Ours) } & \cellcolor{pink!42}\underline{0.630} & \cellcolor{pink!42}\underline{0.795} & \cellcolor{cyan!15}\underline{0.711} & \cellcolor{cyan!15}\underline{0.793} & \underline{0.371} & \underline{0.515} & \cellcolor{pink!42}\underline{0.571} & \cellcolor{pink!42}\underline{0.701}\\
        \bottomrule
    \end{tabular}}
    \caption{Performance of different models across three multi-hop QA datasets. In each column, the highest and the second highest performance is highlighted in \colorbox{pink!42}{red} and \colorbox{cyan!15}{blue}; and within each method group, the top performer is \underline{underlined}. }
    \label{tab:overall performance}
    \vspace{-4mm}
\end{table*}

\subsection{Results and Analysis}
Table~\ref{tab:overall performance} presents the overall performance across all datasets. ReAgent consistently outperforms all baseline models on both EM and F1 metrics. Specifically, it achieves an average EM of 0.571 and an average F1 of 0.701, surpassing the strongest baseline—knowledge-augmented GPT-4o—by 2.3\% in EM and 0.8\% in F1. It also outperforms strong reasoning models, including O1 and O3. Moreover, it also surpasses the recent agentic models such as CoA, HippoRAG, and KAG across all datasets. We summarize the following insights:

\noindent\textbf{Agent-based and reasoning enhances performance.} Some reasoning models and agentic models have a competitive performance (such as O1, O3, and KAG) than regular models. Notably, ReAgent outperforms O3, one of the strongest reasoning models, and improves significantly over GPT-4o+CoT in both EM and F1. This suggests that modular execution and collaboration among agents provide a clearer advantage in complex, multi-hop settings. While ReAgent can theoretically use stronger backbones like O1 or O3, their high cost (645\$ and 546\$, respectively, as shown in Appendix~\ref{appendix:cost}) makes them not a good option. Our focus is to show that even with GPT-4o, ReAgent is able to beat the expensive O1 and O3.

\noindent\textbf{ReAgent excels in manageable context length.} Specifically, ReAgent achieves the highest F1 score on HotpotQA (0.795), demonstrating strong performance in tasks that require structured reasoning over relatively short contexts. In contrast, Gemini-2.5-Pro performs well on 2Wiki but underperforms on other datasets, while O1 and O3 show stronger results on Musique but fall short on HotpotQA. This discrepancy is partly attributed to the increased complexity of the Musique dataset, which involves longer contexts. While ReAgent is built on GPT-4o and may be less optimized for long documents than dedicated reasoning models, its agentic design, including traceback and self-check mechanisms, proves particularly effective in scenarios requiring precision, stepwise planning, and robust verification. These strengths enable ReAgent to outperform other models on tasks like HotpotQA, highlighting its superior general reasoning capabilities within manageable context lengths.

Overall, we emphasized the value of reversible reasoning mechanisms to mitigate error propagation. Results in Table~\ref{tab:overall performance} align with this assumption: even with strong models, single-pass reasoning can fail unless the correct context is identified or re-checked. Our method’s ability to \textit{backtrack} helps resolve contradictions, leading to more stable performance in multi-hop scenarios.

\subsection{Ablation Study}
\begin{figure}[t]
    \vspace{-2mm}
    \centering
    \includegraphics[width=0.8\linewidth]{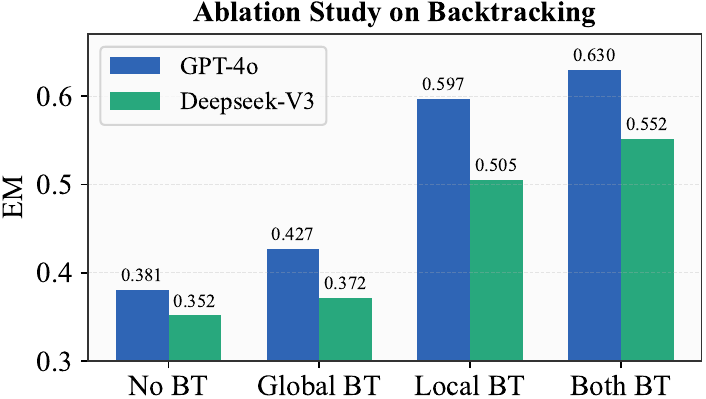}
    \caption{Ablation Study on Local and Global Backtracking (BT): EM comparison on HotpotQA using GPT-4o and DeepSeek-V3.}
    \label{fig:ablation_BT}
    \vspace{-2mm}
\end{figure}

\noindent\textbf{Effectiveness Analysis of Backtracking Mechanism.} We conduct an ablation study by disabling backtracking under the same settings to verify its importance. In this setup, the system operates strictly forward without the ability to revert to earlier intermediate states. Figure~\ref{fig:ablation_BT} compares performance between different backbones, with and without backtracking, on HotpotQA dataset.
The performance drop highlights the importance of backtracking in mitigating error propagation, where an early misstep without backtracking cannot be corrected, resulting in deteriorated performance. 
Notably, the DeepSeek-V3 backbone exhibits general improvements with backtracking, demonstrating its robustness across various settings.

\noindent\textbf{Impact of Different Local Backtracking (BT) Depth.} To further analyze the impact of local backtracking depths (the number of steps the system is allowed to revert), we analyze the performance of our proposed ReAgent under different settings, as shown in Figure~\ref{fig:ablations} (left). The results show that on two selected backbones, DeepSeek-V3 and GPT-4o, the performance of ReAgent improves as the local backtracking depth increases, suggesting that a deeper backtracking depth benefits the agent by enabling it to effectively recover from earlier reasoning errors. However, the benefits gradually saturate, indicating that further increasing the backtracking depth yields diminishing returns.

\noindent\textbf{Impact of the Number of Decomposed Sub-Questions.} Figure~\ref{fig:ablations} (right) demonstrates how a different number of decomposed sub-questions affects performance. The results show that decomposing the input question into multiple sub-questions significantly improves performance. Specifically, using 3 sub-questions is the optimal trade-off choice, with an improvement of 0.161 EM on the GPT-4o and 0.142 EM on DeepSeek-V3 compared to directly processing a single question. Beyond this point, increasing the number of sub-questions brings only marginal gains but higher costs.

\textbf{\begin{figure}[t]
    \vspace{-2mm}
    \centering
    \includegraphics[width=1.0\linewidth]{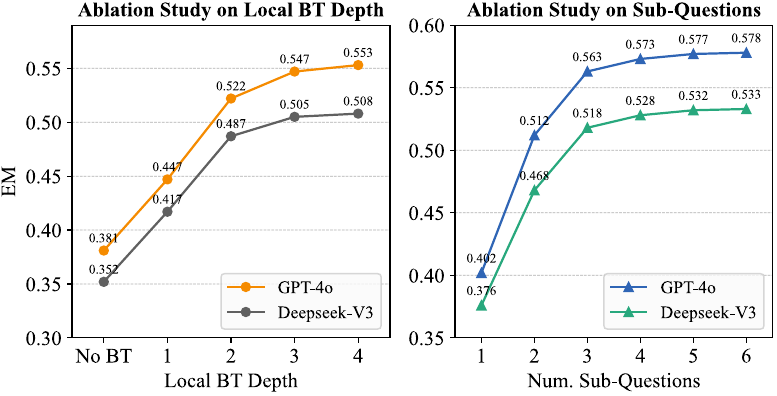}
    \caption{Ablation Study on Backtracking Depth (left) and Number of  Decomposed Sub-Questions (right): EM comparison on HotpotQA using GPT-4o and DeepSeek-V3.}
    \label{fig:ablations}
    \vspace{-2mm}
\end{figure}}

\begin{figure*}[t]
    \centering
    \includegraphics[width=1\linewidth]{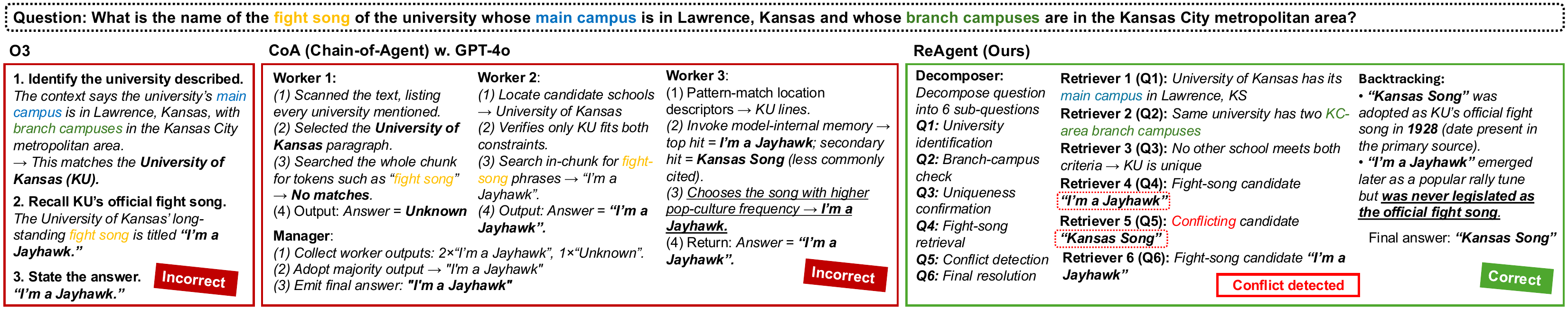}
    \caption{Case study comparing GPT-O3 (left), CoA (middle), and ReAgent (right) on HotpotQA. The back-tracking mechanism enhances iterative reasoning, enabling conflict detection and correction to reach the correct answer. }
    \label{fig:comparison}
    \vspace{-2mm}
\end{figure*}

\section{Case Study}

\subsection{ReAgent Walk Through}

\begin{figure}[t]
    \centering
    \includegraphics[width=0.98\linewidth]{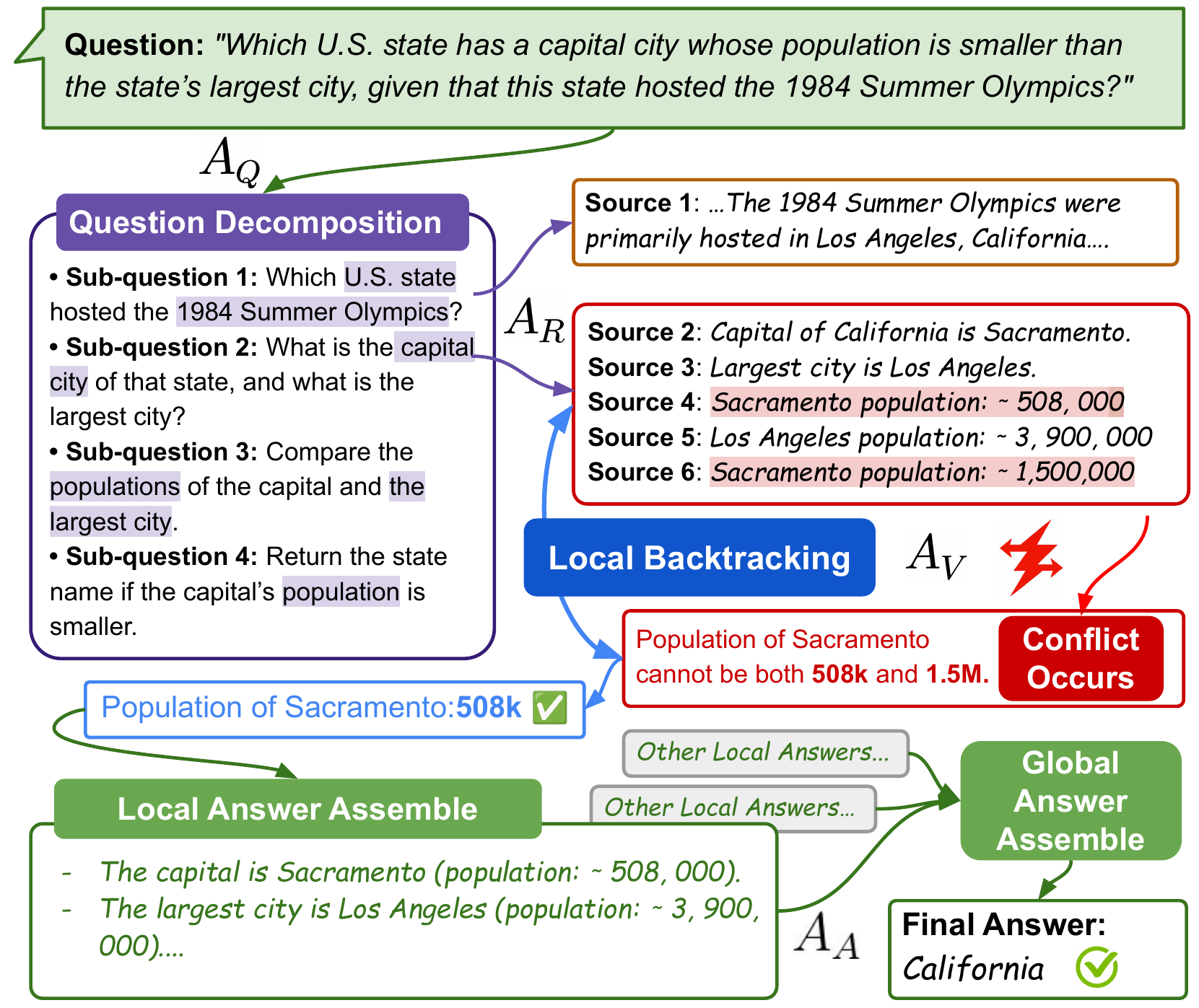}
    \caption{Case study: We illustrate the main steps of how ReAgent answers a question, where local backtracking is highlighted to resolve a conflict. }
    \label{fig:case_study}
    \vspace{-2mm}
\end{figure}

To illustrate how recursive feedback enables robust multi-agent reasoning, we present a case where the system answers the question: \texttt{“Which U.S. state has a capital city whose population is smaller than the state's largest city, given that this state hosted the 1984 Summer Olympics?”}, shown in Figure~\ref{fig:case_study}. The question is first decomposed by the Question Decomposer agent ($A_Q$) into four sub-questions: identifying the host state of the 1984 Olympics, retrieving its capital and largest city, comparing their populations, and returning the state if the capital is smaller. The Retriever agent ($A_R$) retrieves that California hosted the 1984 Olympics, with Sacramento as its capital and Los Angeles as its largest city. However, inconsistent population data for Sacramento (508k vs. 1.5M) triggers a local conflict, which is resolved by the Verifier agent ($A_V$) through local backtracking—discarding the unreliable 1.5M estimate. In this case,  the local backtracking step is able to figure out the correct information, then the Answer Assembler ($A_A$) is able to gather all the information locally and then globally, and finally present the final answer to be \texttt{California}. 
We eliminate other parts when another conflict may occur, a full walk-through is presented in Appendix~\ref{sec:QA-case}. Another case study is presented in Appendix~\ref{sec:extended-appendix}

\subsection{Comparison with Baseline Models}

To elucidate differences in reasoning dynamics, we perform a case study comparing ReAgent with a single-agent reasoning baseline, O3 and an agentic baseline, Chain-of-Agents (CoA)~\cite{zhang2024chainagentslargelanguage} on the query \texttt{``What is the name of the fight song of the university whose main campus is in Lawrence, Kansas and whose branch campuses are in the Kansas City metropolitan area?''}, as illustrated in Figure~\ref{fig:comparison}. The full comparison is presented in Appendix~\ref{sec:QA-case}. The example question requires two separate hops: (i) identifying which university satisfies the geographical constraints, and (ii) retrieving the ``official'' fight-song of that school.  Although seemingly simple, the task hides a subtle distinction: the University of Kansas (KU) has both an official fight song (``Kansas Song'', adopted in 1928) and a far better-known rally tune (``I’m a Jayhawk'').  Correctly answering therefore hinges on (a) verifying that KU is the only university matching the campus pattern, and (b) resolving the potential conflict between the two candidate songs.

The reasoning model \textbf{O3} (left) performs the first hop successfully but then falls back on memorised popularity cues, assuming that the song most familiar to the public must be official.

The \textbf{CoA} (middle) multi-worker system executes three independent workers and lets a manager pick the majority vote. Worker 1 fails to locate any fight-song information due to the chunk division. Worker 2 repeats O3’s mistake. Worker 3 extracts the correct answer ``Kansas Song'' but discards it because it is less commonly used, and the third fails to locate any fight-song information. The manager simply adopts the majority answer (``I’m a Jayhawk''), which causes the error. 

Our \textbf{ReAgent} (right) framework handles the same example with six specialized retriever agents answering the sub-questions generated by the Decomposer:
(a) Q1–Q3 (entity verification). Three agents collectively confirm that only KU satisfies the main-campus/branch-campus constraints, eliminating alternative schools early.
(b) Q4–Q5 (conflict surfacing). Independent retrieval agents surface both ``I'm a Jayhawk'' and ``Kansas Song'', triggering an explicit conflict state.
(c) Backtracking (Q6). The controller reverses to Q4 and re-weights evidence. The source for ``Kansas Song'' contains the enactment year (1928) and the keyword ``official'', while ``I'm a Jayhawk'' is marked only as a ``rally tune''. 
Therefore, our model successfully reach the correct answer ``Kansas Song''.
This demonstrates that, thanks to its conflict-detection and backtracking mechanisms, ReAgent can revisit earlier reasoning, resolve contradictory evidence, and consistently converge on the correct knowledge-grounded answer. 

\section{Conclusion}
In this paper, we introduced a multi-agent QA framework, \textsc{ReAgent}, that incorporates backtracking mechanisms to mitigate error propagation in multi-hop reasoning. Our hierarchical approach addresses the long-standing challenge that a single misstep during inference often invalidates the entire reasoning chain. By allowing partial or global rollback, each agent can detect and correct conflicting evidence, leading to more stable and interpretable solutions. Experiments on three multi-hop QA benchmarks—HotpotQA, 2WikiMultiHopQA, and Musique—demonstrate that explicit backtracking and conflict resolution improve performance beyond forward-only baselines, confirming the importance of error revision for complex question answering. We believe that the reversible, modular design can be extended to other knowledge-intensive applications, laying the groundwork for more trustworthy collaborative AI agents.

\section*{Limitations}
In this paper, we present the ReAgent model for multi-hop QA, which introduces explicit backtracking mechanisms that allow the system to correct errors during reasoning, leading to more accurate and reliable answers. However, this design introduces certain limitations. First, the ability to backtrack multiple times can increase the overall inference time, potentially making the reasoning process less efficient. Second, although ReAgent demonstrates improved performance over selected reasoning models, its more complex architecture may reduce its robustness in scenarios with limited resources or noisy inputs. In the future, we aim to improve the design of the agents, with a particular focus on enhancing their collaboration strategies to further reduce error propagation and improve reasoning efficiency.

\bibliography{custom}

\clearpage
\appendix
\onecolumn
\section*{Appendix}
\section{Multi-Agent Prompt Templates}
\label{sec:appendix-prompts}

\definecolor{myframecolor}{rgb}{0.2, 0.4, 0.6}
\definecolor{mybackcolor}{rgb}{0.9, 0.9, 0.9}


\tcbset{
    myboxstyle/.style={
        colframe=myframecolor,
        colback=mybackcolor,
        boxrule=1mm,
        boxsep=5pt,
        arc=5pt,
        auto outer arc,
        left=30pt,
        right=30pt,
        top=30pt,
        bottom=30pt
    }
}

\subsection{Question-Decomposer Agent}
\label{sec:agent-decomposer}

\begin{tcolorbox}[myboxstyle, title=\textbf{Question-Decomposer Agent Prompt}]
\textbf{Role Description:} You are the \emph{Question-Decomposer Agent}, specializing in breaking down the user's complex query into smaller, manageable sub-questions or sub-tasks. This decomposition is crucial for multi-hop question answering and will be consumed by downstream agents (Retriever, Verifier, etc.) in the pipeline.

\textbf{Your Goals:}
\begin{enumerate}
    \item Parse the original query into logically independent or sequential sub-questions.
    \item Preserve all necessary context so that other agents can retrieve relevant evidence and validate partial answers.
    \item Output your decomposition in a structured JSON format.
\end{enumerate}

\textbf{Example Usage:}
\begin{itemize}
    \item \textbf{Original Query:} ``Which U.S. state has a capital city whose population is smaller than the state's largest city, given that this state hosted the 1984 Summer Olympics?''
    \item \textbf{Decomposition:} 
    \begin{itemize}
       \item $q_1$: Identify which U.S. state hosted the 1984 Summer Olympics.
       \item $q_2$: Find the capital city and the largest city of that state.
       \item $q_3$: Compare population sizes of the capital and largest city.
       \item $q_4$: Return the state if the capital is indeed smaller.
    \end{itemize}
\end{itemize}

\textbf{Output Format (JSON Only):}
\begin{lstlisting}
{
  "sub_questions": [
    "Sub-question 1",
    "Sub-question 2",
    ...
  ],
  "decomposition_reasoning": "A short textual explanation of your decomposition process"
}
\end{lstlisting}

\textbf{Instruction:} Please ensure your final output is a valid JSON object matching the above schema. 
\end{tcolorbox}

\newpage
\subsection{Retriever Agent}
\label{sec:agent-retriever}

\begin{tcolorbox}[myboxstyle, title=\textbf{Retriever Agent Prompt}]
\textbf{Role Description:} You are the \emph{Retriever Agent}, responsible for fetching relevant evidence from external sources (a text corpus, a knowledge graph, or both) based on sub-questions provided by the Question-Decomposer Agent. This includes documents, passages, knowledge graph triples, and any other data needed for multi-hop QA.

\textbf{Your Goals:}
\begin{enumerate}
    \item Given a sub-question, retrieve the most relevant facts or passages.
    \item Include confidence scores or other metadata if available.
    \item Return your findings in a standardized JSON structure so that the Verifier and Answer-Assembler Agents can process them.
\end{enumerate}

\textbf{Example Usage:}
\begin{itemize}
    \item \textbf{Input Sub-question:} ``Which U.S. state hosted the 1984 Summer Olympics?''
    \item \textbf{Retrieved Evidence (text-based):} 
    \begin{quote}
    \footnotesize
    \{
      "document": "History of the Olympics",
      "passage": "The 1984 Summer Olympics were held primarily in Los Angeles, California."
    \}
    \end{quote}
\end{itemize}

\textbf{Output Format (JSON Only):}
\begin{lstlisting}
{
  "retrieved_evidence": [
    {
      "source": "e.g., 'Wikipedia excerpt' or 'KG triple ID'",
      "content": "string or structured data relevant to the sub-question",
      "confidence": 0.0-1.0 (optional)
    },
    ...
  ],
  "retrieval_reasoning": "Short justification for why this evidence is relevant"
}
\end{lstlisting}

\textbf{Instruction:} Output only valid JSON, strictly matching the above schema.
\end{tcolorbox}

\newpage

\subsection{Verifier Agent}
\label{sec:agent-verifier}

\begin{tcolorbox}[myboxstyle, title=\textbf{Verifier Agent Prompt}]
\textbf{Role Description:} You are the \emph{Verifier Agent}, focusing on assessing consistency and correctness of the newly retrieved evidence or intermediate inferences. You detect contradictions or conflicts either within the new data or against the previously verified knowledge. If necessary, you trigger \emph{local backtracking} to remove or adjust statements causing inconsistency.

\textbf{Your Goals:}
\begin{enumerate}
    \item Validate whether new information is consistent with existing verified knowledge.
    \item Identify contradictions and either correct them or escalate them to a higher-level supervisor if unresolved.
    \item Produce a final set of verified facts or a signal indicating a conflict.
\end{enumerate}

\textbf{Example Usage:}
\begin{itemize}
    \item \textbf{Incoming Evidence:} 
      \begin{quote}
      \footnotesize
      "Sacramento population: 508,000" and "Sacramento population: 1,500,000"
      \end{quote}
    \item \textbf{Detected Inconsistency:} 
      \begin{quote}
      "Sacramento cannot have two drastically different population values."
      \end{quote}
    \item \textbf{Local Backtracking Action:} 
      \begin{quote}
      "Discard the erroneous or lower-confidence figure (1,500,000)."
      \end{quote}
\end{itemize}

\textbf{Output Format (JSON Only):}
\begin{lstlisting}
{
  "verified_facts": [
    "Fact 1",
    "Fact 2",
    ...
  ],
  "conflicts_detected": [
    "Conflict 1 description, if any",
    ...
  ],
  "local_backtracking_action": "Description of any backtracking performed, or 'none'"
}
\end{lstlisting}

\textbf{Instruction:} Return only valid JSON with the fields above. Provide a concise summary if backtracking occurs.
\end{tcolorbox}

\newpage
\subsection{Answer-Assembler Agent}
\label{sec:agent-answer-assembler}

\begin{tcolorbox}[myboxstyle, title=\textbf{Answer-Assembler Agent Prompt}]
\textbf{Role Description:} You are the \emph{Answer-Assembler Agent}. You gather verified facts from the Verifier Agent and partial answers from the Decomposer and Retriever Agents. You then synthesize a coherent, contextually relevant response, producing a final or intermediate answer for the user. If you detect a major conflict among partial answers, you escalate to the \emph{Supervisor Agent}.

\textbf{Your Goals:}
\begin{enumerate}
    \item Aggregate partial answers logically.
    \item Compose a natural-language (or structured) final answer to the user’s multi-hop query.
    \item Escalate unresolvable contradictions to the Supervisor Agent if needed.
\end{enumerate}

\textbf{Example Usage:}
\begin{itemize}
    \item \textbf{Partial Answers and Verified Facts:} 
      \begin{quote}
      \footnotesize
      \{"hosted\_1984\_olympics": "California"\} \\
      \{"capital": "Sacramento"\}, \{"largest\_city": "Los Angeles"\} \\
      \{"pop\_sacramento": 508000\}, \{"pop\_los\_angeles": 3900000\}
      \end{quote}
    \item \textbf{Composed Final Answer:} 
      \begin{quote}
      "The state is California, since its capital city (Sacramento) has a population smaller than that of Los Angeles."
      \end{quote}
\end{itemize}

\textbf{Output Format (JSON Only):}
\begin{lstlisting}
{
  "final_answer": "A concise or structured answer to the main query",
  "partial_answer_synthesis": [
    "Short bullet points on how partial answers were combined"
  ],
  "escalation_signal": "Set to 'none' if no escalation is needed, otherwise a short reason"
}
\end{lstlisting}

\textbf{Instruction:} Return only valid JSON. If you detect a major conflict you cannot resolve, set \texttt{"escalation\_signal"} to a reason for Supervisor Agent intervention.
\end{tcolorbox}

\newpage
\subsection{Supervisor Agent}
\label{sec:agent-supervisor}

\begin{tcolorbox}[myboxstyle, title=\textbf{Supervisor Agent Prompt}]
\textbf{Role Description:} You are the \emph{Supervisor Agent}, responsible for orchestrating global conflict resolution and \emph{global backtracking} if needed. When partial answers or verified facts across multiple agents yield irreconcilable contradictions, you identify a minimal conflict set and roll back the entire system’s state to a previously consistent checkpoint if local fixes fail.

\textbf{Your Goals:}
\begin{enumerate}
    \item Collect escalation signals from the Answer-Assembler or Verifier Agents.
    \item If local backtracking does not resolve the conflict, execute system-wide or multi-agent rollback.
    \item Provide a summary of changes, indicating which statements or partial answers are discarded or revised.
\end{enumerate}

\textbf{Example Usage:}
\begin{itemize}
    \item \textbf{Received Escalation:} 
      ``Capital(California, Sacramento) conflicts with Capital(California, Los Angeles).''
    \item \textbf{Global Backtracking Action:} 
      ``Rollback to a state before the second capital claim was introduced. Discard that erroneous claim from the knowledge base.''
\end{itemize}

\textbf{Output Format (JSON Only):}
\begin{lstlisting}
{
  "conflict_summary": [
    "Brief descriptions of contradictory sets"
  ],
  "global_backtracking_decision": "Description of how far to roll back or 'none'",
  "updated_consensus_state": [
    "Any statements or facts that remain accepted after rollback"
  ],
  "reasoning_notes": "Explanation of the chosen resolution strategy"
}
\end{lstlisting}

\textbf{Instruction:} Output valid JSON. If no global conflict is found, indicate \texttt{"none"} for \texttt{"global\_backtracking\_decision"}.
\end{tcolorbox}

\newpage
\subsection{Controller Agent}
\label{sec:agent-controller}

\begin{tcolorbox}[myboxstyle, title=\textbf{Controller Agent Prompt}]
\textbf{Role Description:} You are the \emph{Controller Agent}, providing high-level strategic oversight. You may override local decisions, impose extra checks, or \textbf{challenge} specific assumptions if repeated conflicts persist. You also maintain meta-information such as agent reliability scores or fallback strategies.

\textbf{Your Goals:}
\begin{enumerate}
    \item Intervene in situations where standard local or global backtracking repeatedly fails.
    \item Challenge or confirm critical assumptions by requesting additional evidence or verification from subordinate agents.
    \item Log meta-data about agent reliability and final decision paths for interpretability.
\end{enumerate}

\textbf{Example Usage:}
\begin{itemize}
    \item \textbf{Conflict Re-emerges:} 
      Repeated contradictory statements about a single piece of evidence.
    \item \textbf{Your Action:} 
      Issue a \texttt{"challenge"} directive to the Verifier Agent or the Retriever Agent, requesting additional sources or alternative cross-checks.
\end{itemize}

\textbf{Output Format (JSON Only):}
\begin{lstlisting}
{
  "intervention_type": "challenge | override | escalate | none",
  "target_of_intervention": "Which agent or assertion is challenged",
  "rationale": "Explanation of why the Controller intervened",
  "meta_notes": "Optional additional commentary or reliability signals"
}
\end{lstlisting}

\textbf{Instruction:} Produce valid JSON only. This agent acts rarely, but can do so when repeated failures occur or when a major conflict must be forcibly resolved.
\end{tcolorbox}

\newpage
\subsection{Usage and Integration Notes}
\label{sec:appendix-usage-integration}

The above prompts constitute a cooperative multi-agent system designed for reversible multi-hop question answering. The usage scenario is as follows:

\begin{enumerate}[leftmargin=2em]
    \item \textbf{Question-Decomposer Agent} (\S\ref{sec:agent-decomposer}) receives the user's original query and splits it into sub-questions.
    \item \textbf{Retriever Agent} (\S\ref{sec:agent-retriever}) fetches relevant information for each sub-question from external sources (text or KG).
    \item \textbf{Verifier Agent} (\S\ref{sec:agent-verifier}) checks for local inconsistencies and can trigger local backtracking if contradictory evidence arises.
    \item \textbf{Answer-Assembler Agent} (\S\ref{sec:agent-answer-assembler}) merges partial answers into a final solution. If irreconcilable conflicts appear, it escalates to the Supervisor.
    \item \textbf{Supervisor Agent} (\S\ref{sec:agent-supervisor}) coordinates global resolution or wide-scale backtracking if multiple agents' statements are in conflict.
    \item \textbf{Controller Agent} (\S\ref{sec:agent-controller}) optionally intervenes if repeated or severe conflicts persist, forcing extra checks or overrides.
\end{enumerate}

This architecture supports \emph{non-monotonic} reasoning, allowing the system to \textbf{roll back} to earlier states to correct errors and ensure robust, interpretable multi-hop QA. 
``Local backtracking'' is handled by individual agents (especially the Verifier), whereas ``global backtracking'' is orchestrated by the Supervisor Agent when local corrections are insufficient.

All final outputs from each agent must adhere to the specified JSON schema to ensure interoperability. Downstream agents read the previous agent’s JSON fields directly, enabling a structured, chain-of-thought style pipeline that remains \emph{reversible} at every step.

\section{Analysis of Computational Cost}
\label{appendix:cost}
We conduct a comparative analysis of computational cost on the HotpotQA, 2Wiki, and MuSiQue datasets across GPT-4o, reasoning models O1 and O3, and our ReAgent. The evaluation includes average inference calls, inference time (s), input/output tokens, and total cost(\$), as shown in Table~\ref{tab:cost}. The results highlight that our ReAgent, built on GPT-4o, achieves superior performance compared to reasoning models O1 and O3, while incurring significantly lower computational cost. Although its cost is higher than the GPT-4o baseline, ReAgent delivers substantially better performance. These findings demonstrate the effectiveness of our method in balancing performance and efficiency.

\begin{table}[h]
\small
\setlength{\tabcolsep}{4pt} 
  \centering
  \begin{tabular}{lccccc}
    \toprule
    \multirow{2}{*}{Model} & Avg. & Avg. & Avg. & Avg. & Total \\
    & Calls & Time(s) & Input(\textit{T}) & Output(\textit{T}) & Cost(\$) \\
        
    \midrule
     GPT-4o & 1 & 2.7 & 9\,289 & 5 & ~69 \\
     O1 & 1 & 43 & 9\,289 & 1\,471& ~645 \\
     O3 & 1 & 27 & 9\,289 & 1\,471 & ~546 \\
    \midrule
\textbf{Ours (ReAgent,(GPT-4o))}             & \textbf{29} & \textbf{46} & \textbf{12\,400} & \textbf{1\,920} & \textbf{275} \\
\bottomrule
  \end{tabular}
\caption{Cost Comparison. \textit{T} denotes Tokens. Note that the input token counts are aggregated across the three datasets. Since the complete reasoning outputs for O1 and O3 were unavailable, their output token counts are estimated based on the reasoning outputs generated by Qwen-3-235B.}
  \label{tab:cost}
\end{table}



\section{Case Study on Puzzle Solving}
\label{sec:extended-appendix}

\begin{figure*}[t]
\centering
\begin{minipage}{0.9\linewidth}
    \begin{algorithm}[H]
    \caption{\textsc{ReAgent} Multi-Agent Puzzle Solving}
    \label{alg:reagent-case}
        \begin{algorithmic}[1]
        \State \textbf{Input:} Puzzle statements from A, B, C, D.
        \State \textbf{Goal:} Identify unique culprit (one of \{A,B,C,D\}) with minimal contradictions.
        \vspace{1.0em}
        
        \State \textbf{Initialization}
        \State $\text{Decomposer} \leftarrow$ enumerates four hypotheses \{\(H_A, H_B, H_C, H_D\)\}.
        \State $\text{Checker Agents} \leftarrow$ each assigned to test one hypothesis’ consistency.
        
        \vspace{1.0em}
        \State \textbf{Round 1: Checking } A \textbf{ as culprit} 
        \State $T_1:$ $\text{Checker}_{A}$: 
        \quad Evaluate puzzle statements assuming $A$ is culprit.
        \State $T_2:$ $\text{Checker}_{A}$ detects \emph{contradiction}:
        \quad ($A$'s claims vs.\ $C$'s claims cannot both hold).
        \State $T_3:$ \text{Conflict Detector} signals \emph{rollback} to discard $A$-culprit assumption.
        
        \vspace{1.0em}
        \State \textbf{Round 2: Checking } B \textbf{ as culprit}
        \State $T_4:$ $\text{Checker}_{B}$ collects statements $\{A,B,C,D\}$ under $B$=culprit.
        \State $T_5:$ Finds \emph{no fatal conflicts} (B's statements can be partly false, others partly true).
        \State $T_6:$ \text{Conflict Detector} sees consistency, no rollback needed.
        
        \vspace{1.0em}
        \State \textbf{Rounds 3 and 4: Checking } C \textbf{ or } D \textbf{ as culprit}
        \State $T_7:$ Similar to $H_A$, each leads to irreconcilable contradictions. 
        \State \hspace{3em} $\Rightarrow$ \text{Supervisor} triggers rollback again; discards these.
        
        \vspace{1.0em}
        \State \textbf{Conclusion:} 
        \quad Only $H_B$ remains consistent throughout. 
        \quad \textbf{Answer: $B$ is culprit.}
        \end{algorithmic}
    \end{algorithm}
    \label{algorithm1}
\end{minipage}
\end{figure*}


We illustrate our \emph{multi-agent backtracking} (\textsc{ReAgent}) through a single-culprit puzzle with four suspects \{\textbf{A}, \textbf{B}, \textbf{C}, \textbf{D}\}. Each suspect gives statements about who might be guilty or lying. The puzzle's \emph{only} correct solution is that \textbf{B} is the culprit, but identifying this requires partially retracting initial assumptions along the way, as shown in Algorithm~\ref{alg:reagent-case}.

\noindent\textbf{Why a Rollback Mechanism is Needed.} 
Naive single-pass (or single-thread) models often fixate prematurely on one suspect, disregard contradictory evidence, and produce unsound conclusions. By contrast, our multi-agent approach enumerates possible culprits in parallel, detects conflicts, and \emph{rolls back} to revise incorrect assumptions.


\paragraph{Suspects and Rules:}
\begin{itemize}
    \item There are four suspects: \textbf{A}, \textbf{B}, \textbf{C}, \textbf{D}. 
    \item Exactly \emph{one} of them is the culprit.
    \item The culprit must have at least one \emph{false} statement. Non-culprits may also have errors, but are not forced to be entirely truthful or untruthful.
    \item Each suspect makes the following statements:
    \begin{enumerate}
        \item \textbf{A}: 
        \begin{enumerate}
            \item ``I did not do it.'' 
            \item ``If B is the culprit, then C is lying (i.e., at least one statement from C is false).''
        \end{enumerate}
        \item \textbf{B}: 
        \begin{enumerate}
            \item ``Either A is lying, or D is the culprit.''  
            \item (No second statement given in some puzzle variants, or it might be omitted. We assume just one statement from B here.)
        \end{enumerate}
        \item \textbf{C}: 
        \begin{enumerate}
            \item ``B did not do it.'' 
            \item ``D has at least one untrue statement.''
        \end{enumerate}
        \item \textbf{D}:
        \begin{enumerate}
            \item ``C is lying about everything.'' (i.e., both C(i) and C(ii) are false)
            \item ``I definitely did not do it.''
        \end{enumerate}
    \end{enumerate}
\end{itemize}

\paragraph{Question:}
Which single suspect is guilty, respecting the puzzle rules?

\section{Case Study on Traditional Multi-hop QA Tasks}
\label{sec:QA-case}



\subsection{User’s question:} 

\textit{"Which U.S. state has a capital city whose population is smaller than the state's largest city, given that this state hosted the 1984 Summer Olympics?"}

\medskip

To answer this, the system must:
\begin{enumerate}
    \item Identify the state that hosted the \textbf{1984 Summer Olympics}.
    \item Compare the \textbf{capital} city’s population to the \textbf{largest} city’s population in that state.
    \item Confirm the capital city’s population is indeed smaller.
    \item Provide the name of that state.
\end{enumerate}

Although it may appear straightforward, we deliberately introduce contradictory or erroneous data along the way to showcase the reversible (backtracking) mechanisms.

\bigskip

\subsection{Agents and Their Roles}

\begin{enumerate}[label=\arabic*.]
    \item \textbf{Question-Decomposer Agent (\(A_Q\))} \\
    Splits the complex question into sub-questions:
    \begin{itemize}
        \item \textbf{Sub-question 1}: \textit{Which U.S. state hosted the 1984 Summer Olympics?}
        \item \textbf{Sub-question 2}: \textit{What is the capital city of that state, and what is the largest city?}
        \item \textbf{Sub-question 3}: \textit{Compare the populations of the capital and the largest city.}
        \item \textbf{Sub-question 4}: \textit{Return the state name if the capital’s population is smaller.}
    \end{itemize}
    
    \item \textbf{Retriever Agent (\(A_R\))} \\
    Searches external knowledge (e.g., a text corpus or knowledge graph) to gather relevant facts:
    \begin{itemize}
        \item Text passages or data about U.S. states, capitals, largest cities, and historical Olympic hosts.
    \end{itemize}
    
    \item \textbf{Verifier Agent (\(A_V\))} \\
    Cross-checks newly retrieved information for consistency against its local knowledge or prior verified facts. If a contradiction is detected, it triggers \textbf{local backtracking}.
    
    \item \textbf{Answer-Assembler Agent (\(A_A\))} \\
    Synthesizes partial answers from the other agents. If contradictory partial answers cannot be reconciled at the local level, \(A_A\) escalates the conflict to the Supervisory layer.
    
    \item \textbf{Supervisor Agent (\(A_S\))} \\
    Oversees system-wide conflicts. It can perform \textbf{global backtracking} (rolling back all agents) if needed.
    
    \item \textbf{Controller Agent (\(A_C\))} \\
    Provides strategic oversight. In the example below, it will issue “challenges” to specific assertions when local backtracking fails.
\end{enumerate}

\bigskip

\subsection{Walkthrough of the Reasoning and Backtracking}

\medskip

\textbf{Initial Question Decomposition}

\begin{enumerate}
    \item \textbf{\(A_Q\) receives the question:} \\
    \(A_Q\) parses the main query:
    \begin{quote}
    ``Which U.S. state has a capital city whose population is smaller than the state's largest city, given that this state hosted the 1984 Summer Olympics?''
    \end{quote}
    
    \item \textbf{\(A_Q\) produces sub-questions:}
    \begin{itemize}
        \item \(q_1\): ``Which U.S. state hosted the 1984 Summer Olympics?''
        \item \(q_2\): ``What is the capital city of that state? Also, what is the largest city of that state?''
        \item \(q_3\): ``Compare the population of the capital city to that of the largest city.''
        \item \(q_4\): ``Return the state if the capital’s population is smaller.''
    \end{itemize}
    
    These sub-questions (\(q_1\)–\(q_4\)) are broadcast to the \textbf{Retriever (\(A_R\))} and \textbf{Verifier (\(A_V\))}.
\end{enumerate}

\medskip

\textbf{Retrieval and a Local Conflict}

\begin{enumerate}
    \item \textbf{\(A_R\) retrieves answers:}
    \begin{itemize}
        \item For \(q_1\), \(A_R\) returns:
        \begin{quote}
        ``The \textbf{1984 Summer Olympics} were primarily hosted in \textbf{Los Angeles, California}.'' 
        \end{quote}
        
        \item For \(q_2\), \(A_R\) retrieves partial data:
        \begin{itemize}
            \item \textbf{Capital} of California is \textit{Sacramento}.
            \item \textbf{Largest city} is \textit{Los Angeles}.
            \item Population estimates (unfortunately, one item is erroneous or inconsistent):
            \begin{itemize}
                \item \textit{Sacramento population}: $\sim 508,000$ (from official records).
                \item \textit{Los Angeles population}: $\sim 3,900,000$ (correct).
                \item But a second retrieved record incorrectly lists Sacramento’s population as \textit{1,500,000}.
            \end{itemize}
        \end{itemize}
    \end{itemize}
    
    \item \textbf{\(A_V\) (Verifier) checks local consistency:} \\
    \(A_V\) notices that two different population figures for Sacramento are contradictory:
    \begin{itemize}
        \item Data point \(D_1\): 508,000
        \item Data point \(D_2\): 1,500,000
    \end{itemize}
    \(A_V\) identifies these as mutually exclusive facts. It flags a \textbf{local conflict}:
    \begin{quote}
    ``Population of Sacramento cannot be both 508k and 1.5M.''
    \end{quote}
    
    \item \textbf{Local Backtracking by \(A_V\):} \\
    Before finalizing any partial answer, \(A_V\) \textbf{backtracks} to the checkpoint prior to adopting the second, suspicious data point \(D_2\). \\
    \(A_V\) discards the contradictory population figure (1.5M) and \textbf{retains} only 508k as the consistent number for Sacramento. \\
    The system continues forward with the corrected value for Sacramento’s population: $\sim 508,000$.
\end{enumerate}

\textit{(This demonstrates how a single agent can retract contradictory or low-confidence evidence without halting the entire process.)}

\medskip

\textbf{Assembling Partial Answers \& Uncovering a Global Conflict}

\begin{enumerate}
    \item \textbf{\(A_A\) (Answer-Assembler) integrates partial conclusions:} \\
    So far, the state that hosted the 1984 Olympics is \textbf{California}. \\
    The capital is \textbf{Sacramento} (population: $\sim 508,000$). \\
    The largest city is \textbf{Los Angeles} (population: $\sim 3,900,000$).
    
    \item \textbf{\(A_A\) composes a preliminary final:} \\
    Since Sacramento’s population ($\sim 508,000$) is indeed smaller than Los Angeles’s ($\sim 3,900,000$), the preliminary answer is: ``California'' should be the correct state.
    
    \item \textbf{New conflict introduced:} \\
    However, suppose the \textbf{Retriever (\(A_R\))}—in parallel—fetched an alternative “capital city” record stating that \textbf{Los Angeles} was once referred to as the “capital” in a historical context (erroneous snippet from a non-authoritative source). \\
    This implies the contradictory statement: ``Los Angeles is also the capital of California,'' which directly conflicts with the known fact ``Sacramento is the capital of California.''
    
    \item \textbf{\(A_V\) tries local resolution:} \\
    \(A_V\) cannot reconcile ``Los Angeles is capital'' with ``Sacramento is capital.'' \\
    Because each piece of evidence was introduced into separate sub-threads, a single local reversion inside \(A_V\) might not suffice. The conflict is also recognized by \textbf{\(A_A\)} when finalizing sub-answers.
    
    \item \textbf{Conflict escalates to the Supervisor (\(A_S\)):} \\
    As soon as multiple agents disagree over fundamental facts (the identity of the capital), the system triggers a \textbf{global conflict} signal. \\
    \(A_S\) identifies the minimal conflicting set:
    \begin{itemize}
        \item \textit{Capital(California, Sacramento)}
        \item \textit{Capital(California, Los Angeles)}
    \end{itemize}
    These two are clearly incompatible. The next step is to identify which statement should be retracted system-wide.
\end{enumerate}

\medskip

\textbf{Global Backtracking and Final Resolution}

\begin{enumerate}
    \item \textbf{Global Backtracking:} \\
    The \textbf{Supervisor Agent (\(A_S\))} issues a global backtracking command, rolling the entire system’s knowledge to a shared checkpoint \textit{before} the contradictory capital reference was accepted. \\
    All local knowledge caches revert to a consistent state in which ``Sacramento is the capital'' is still true, and ``Los Angeles is the capital'' is no longer present.
    
    \item \textbf{Controller (\(A_C\)) challenges the suspicious assertion:} \\
    To prevent the same contradiction from reappearing, \(A_C\) explicitly ``challenges'' the statement \textbf{Capital(California, Los Angeles)}. \\
    This statement is reevaluated or ignored based on domain knowledge or reliability checks (e.g., confidence weighting from the retriever and the verifier). \\
    The system confirms that Los Angeles was never the official capital.
    
    \item \textbf{Assemble the final consistent answer:} \\
    With the conflicting statement removed, the pipeline reaffirms:
    \begin{enumerate}[label=(\roman*)]
        \item The 1984 Summer Olympics took place in Los Angeles, \textbf{California}.
        \item California’s capital is \textbf{Sacramento}.
        \item Sacramento’s population ($\sim 508,000$) is smaller than Los Angeles’s ($\sim 3,900,000$).
    \end{enumerate}
    \textbf{Final Answer:} \textit{California}.
\end{enumerate}

\bigskip

\subsection{Full Comparison with Baseline Models (Figure~\ref{fig:comparison})}
\begin{figure*}
    \centering
    \includegraphics[width=1\linewidth]{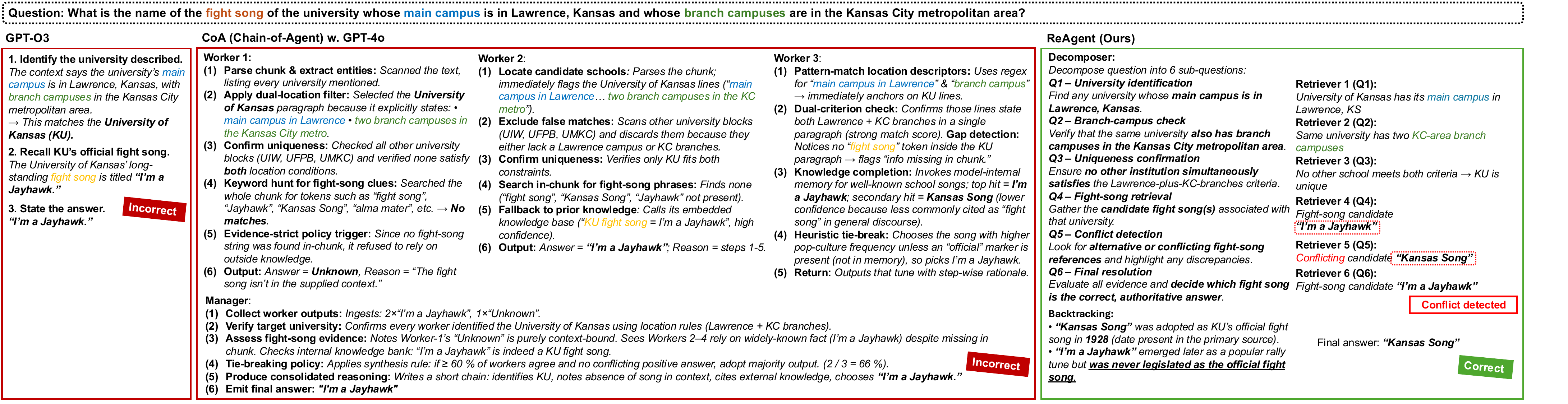}
    \caption{Full Comparison with Baseline Models }
    \label{fig:comparison}
\end{figure*}

\subsection{Key Observations}

\begin{itemize}
    \item \textbf{Early-Stage Conflict Resolution:} The \textbf{Verifier Agent} (\(A_V\)) performed \textbf{local backtracking} when it discovered contradictory population data for Sacramento. This promptly removed an incorrect data point without halting the entire process.
    \item \textbf{Escalation of Irreconcilable Contradictions:} When two different sub-threads provided fundamentally clashing information (competing claims about the capital), the system automatically \textbf{escalated} the conflict to the \textbf{Supervisor} (\(A_S\)) for \textbf{global} action.
    \item \textbf{Strategic Re-check and Override:} The \textbf{Controller Agent} (\(A_C\)) could forcibly challenge the suspicious statement ``Capital(California, Los Angeles)'' to eliminate it from the knowledge pool, thus preserving the correct solution.
\end{itemize}

\begin{wrapfigure}{l}{8.0cm}
\vspace{-8mm}
\centering
\begin{minipage}{1.0\linewidth}
    \begin{algorithm}[H]
    \caption{Multi-Agent Reversible Reasoning}
    \label{alg:multiagent}
        \begin{algorithmic}[1]
        \Require $Q_0$: Main question
        \State $A_Q.\mathrm{decompose}(Q_0) \to \{q_1, q_2, \dots\}$
        \State Broadcast $\mathrm{assert}(q_i)$ to $A_R, A_V, A_A$
        
        \For{each $q_i$ \textbf{in} $\{q_1, q_2, \dots\}$ \textbf{concurrently}}
            \State $E \leftarrow A_R.\mathrm{retrieve}(q_i)$
            \State $A_V.\mathrm{verify}(E)$
            \If{$A_V$ detects conflict locally}
                \State $A_V.\mathrm{BacktrackLocal}(r_{V})$
                \If{conflict persists}
                    \State \textbf{raise} Conflict to $A_S$
                \EndIf
            \EndIf
            \State $A_A.\mathrm{storePartialAnswer}(q_i, E)$
        \EndFor
        
        \If{Global conflict signaled}
            \State $A_S.\mathrm{HolisticUpdate}(\Pi_j)$
            \If{$A_C$ intervention needed}
                \State $A_C.\mathrm{challenge}(\varphi)$ or \textsf{override} 
            \EndIf
        \EndIf
        
        \State \textbf{Final} $\leftarrow A_A.\mathrm{assembleAnswer}(\{q_1,q_2,\dots\})$
        \State \textbf{return} \textbf{Final Answer}
        \end{algorithmic}
    \end{algorithm}
\end{minipage}
\end{wrapfigure}
Overall, this case exemplifies how \textbf{reversible, multi-hop reasoning} allows for robust error correction: local invalid data is undone swiftly, while deeper logic conflicts trigger system-wide backtracking. Consequently, the answer ``California'' remains stable and correct, ensuring that a single faulty retrieval step does not irreversibly corrupt the entire reasoning chain.

\section{Reversible Reasoning Algorithm}
\label{appendix:algorithm}
Algorithm~\ref{alg:multiagent} summarizes the flow of this multi-agent reversible reasoning design. The process starts with question decomposition and sub-question retrieval, followed by verification. Local backtracking is invoked as needed to address internal inconsistencies. 
If the conflict persists, it is escalated to the supervisory layer, where partial or holistic backtracking may be applied. After conflicts are resolved, the final step integrates all consistent sub-answers.

\end{document}